%% file: 00-cb2demo.tex
\pdfoutput=1

\documentclass[11pt]{article}

\usepackage{ACL2023}

\usepackage{times}
\usepackage{latexsym}

\usepackage[T1]{fontenc}

\usepackage[utf8]{inputenc}

\usepackage{microtype}

\usepackage{inconsolata}

\input{00-header.tex}

\title{\gamename: Collaborative Natural Language Interaction Research Platform}

\author{Jacob Sharf, Mustafa Omer Gul, \and Yoav  Artzi\\
  Department of Computer Science and Cornell Tech, Cornell University  \\  \texttt{jacobsharf@gmail.com} \hspace{10pt} \texttt{\{momergul, yoav\}@cs.cornell.edu}}

\begin{document}
\maketitle

\begin{abstract}
\input{01-abstract.tex}

\end{abstract}

\input{10-intro}

\input{20-related}

\input{30-design}

\input{40-impl}

\input{50-tasks}

\input{60-crowdsourcing}

\input{70-data}

\input{80-conclusion}

\input{90-acks}

\input{95-ethics}

\bibliography{main_local_ver,local}
\bibliographystyle{acl_natbib}

\end{document}

%% file: 00-header.tex
\usepackage{graphicx}
\usepackage{xspace}

\usepackage{caption}
\usepackage{subcaption}

\newcommand{\aautoref}[1]{\hyperref[#1]{Appendix~\ref*{#1}}}

\usepackage{graphicx}

\usepackage{array}

\usepackage{tabularx}

\usepackage[frozencache,cachedir=.]{minted}
\usepackage{tikz}

\usepackage[inkscapearea=page]{svg}

\usepackage{adjustbox}

\usepackage{float}

\usepackage{booktabs}

\newcolumntype{S}{>{\hsize=.5\hsize}X}
\newcolumntype{M}{>{\hsize=.333\hsize}X}
\newcolumntype{L}{>{\hsize=2.5\hsize}X}

\newcolumntype{U}{>{\hsize=\hsize}X}
\newcolumntype{D}{>{\hsize=2\hsize}X}

\usepackage{titlesec}
\titlespacing*{\paragraph} {0pt}{.8ex plus .1ex minus .1ex}{1em}

\newcommand{\eat}[1]{}

\newcommand{\cerealbar}{\textsc{CerealBar}\xspace}
\newcommand{\gamename}{CB2\xspace}

%% file: 01-abstract.tex
\gamename is a multi-agent platform to study collaborative natural language interaction in a grounded task-oriented scenario. It includes a 3D game environment, a backend server designed to serve trained models to human agents, and various tools and processes to enable scalable studies. We deploy \gamename at \url{https://cb2.ai} as a system demonstration with a learned instruction following model.

%% file: 10-intro.tex
\section{Introduction}\label{sec:intro}

Collaborative grounded natural language interactions involve multiple agents, either human or machine, working together to complete tasks while coordinating using natural language.
A key obstacle in studying such scenarios is building the research interaction platform,  a significant design and engineering undertaking. 
This requires building and designing the interaction environment, the task the agents collaborate on, an interface for both machine learning models and human agents, and a process to onboard human agents. 
Each aspect dramatically influences the interaction and language elicited, and is critical to get right.

We introduce \gamename, a platform for the study of collaborative grounded natural language interaction, and demonstrate its use through the deployment of a learned collaborative natural language agent. 
\gamename largely instantiates the \cerealbar scenario~\citep{Suhr2019:cerealbar},\footnote{\gamename introduces several optional modifications to \cerealbar aimed at richer language and tighter collaboration.} but is implemented from scratch to emphasize research accessibility. 
\gamename is a customizable, scalable, and complete research platform, including server and clients for multi-agent human-machine interactions, tools for real-time data management, and processes to onboard crowdsourcing workers.

The \gamename scenario poses learning and reasoning challenges, as well as opportunities. 
Comprehending and producing instructions in \gamename requires addressing the symbol grounding problem~\cite{Harnad1990:symbol-grounding-problem}, which is studied extensively in the instruction following~\cite[e.g.,][]{Chen:11, Artzi:13,Misra:17instructions,Fried:17pragmatic-models} and generation~\cite[e.g.,][]{Mei:16generation,Wang2021:generatingInstructions} literature. 
However, the collaborative scenario remains relatively understudied. 
Collaboration is not simply an added complication, but dramatically alters both interaction and learning through joint presence and action. 
It allows the instructor to ad-hoc modify the tasks they delegate based on the follower behavior, potentially recovering from system failures. 
At the same time, this adaptation creates constant distribution shift, a significant generalization challenge. 
Learning is also drastically transformed through collaboration.
The constant engagement of other agents (including humans), the ability to modify delegation strategies, and the shared task-based incentives  bring about within-interaction signals that can be used for continual learning, reducing the dependency on annotated data and enabling model adaptation.

We deploy a demonstration of \gamename with a learned baseline instruction following agent (\autoref{sec:deployment}). 
Players can connect to \gamename and collaborate with our agent or other human agents at \url{https://cb2.ai/}.\footnote{Our deployment has received IRB exemption. All recorded data is anonymized.}
The \gamename platform is available at \url{https://github.com/lil-lab/cb2}.
A video demonstration of \gamename is available at \url{https://youtu.be/tALpX_KKmIw}.

%% file: 20-related.tex
\section{Related Work}\label{sec:related}

\gamename is a re-implementation and extension of 
\cerealbar, a scalable platform to study natural language instruction collaboration~\cite{Suhr2019:cerealbar}. 
\cerealbar was used to study  instruction following~\cite{Suhr2019:cerealbar,Suhr2022:continualfollowing}, instruction generation~\cite{Kojima2021:gen-learn}, and linguistic change~\cite{Effenberger2021:cerealbar-analysis}. 

\gamename is related to instruction following environments, such as SAIL~\cite{MacMahon:06}, R2R~\cite{Anderson:18r2r}, RxR~\cite{Ku2020:room-across-room}, and ALFRED~\cite{Shridhar2020:alfred}. 
In contrast, \gamename is focused on embodied multi-agent collaborations, including with human agents.  

Symbol grounding~\cite{Harnad1990:symbol-grounding-problem}, a core challenge in \gamename, was studied extensively in the single-agent context of instruction following~\cite[e.g.,][]{Chen:11, Artzi:13, Fried:17pragmatic-models, Blukis:18drone} and generation~\cite[e.g.,][]{daniele2016natural, Kojima2021:gen-learn, Wang2021:generatingInstructions}.
The \gamename scenario emphasizes multi-agent collaboration, an aspect that is significantly less studied with natural language instruction. 
The Cards corpus~\cite{Djalali12:cards-preference, Potts:12} presents a related scenario, which has been used for linguistic analysis. 
A related problem is studied by the emergent communication literature~\cite{Lazaridou:17, Andreas:17, Lazaridou2020:emegent-lang-survey}, but with less focus on collaboration with human agents. 
Natural language collaboration between agents with asymmetric capabilities has also been studied with Minecraft-based scenarios~\cite{Narayan2019:collaborative-minecraft-diaglogue, Jayannavar2020:instructions-minecraft-dialogue, kiseleva2022iglu}. 
\gamename differs from these in allowing both agents to effect changes on the environment, enabling ad-hoc modification and delegation of tasks.

%% file: 30-design.tex
\section{Interaction Scenario}\label{sec:scenario}

\begin{figure*}[h!]
    \centering
    \begin{subfigure}[b]{\textwidth}
        \frame{\includegraphics[width=\linewidth,keepaspectratio]{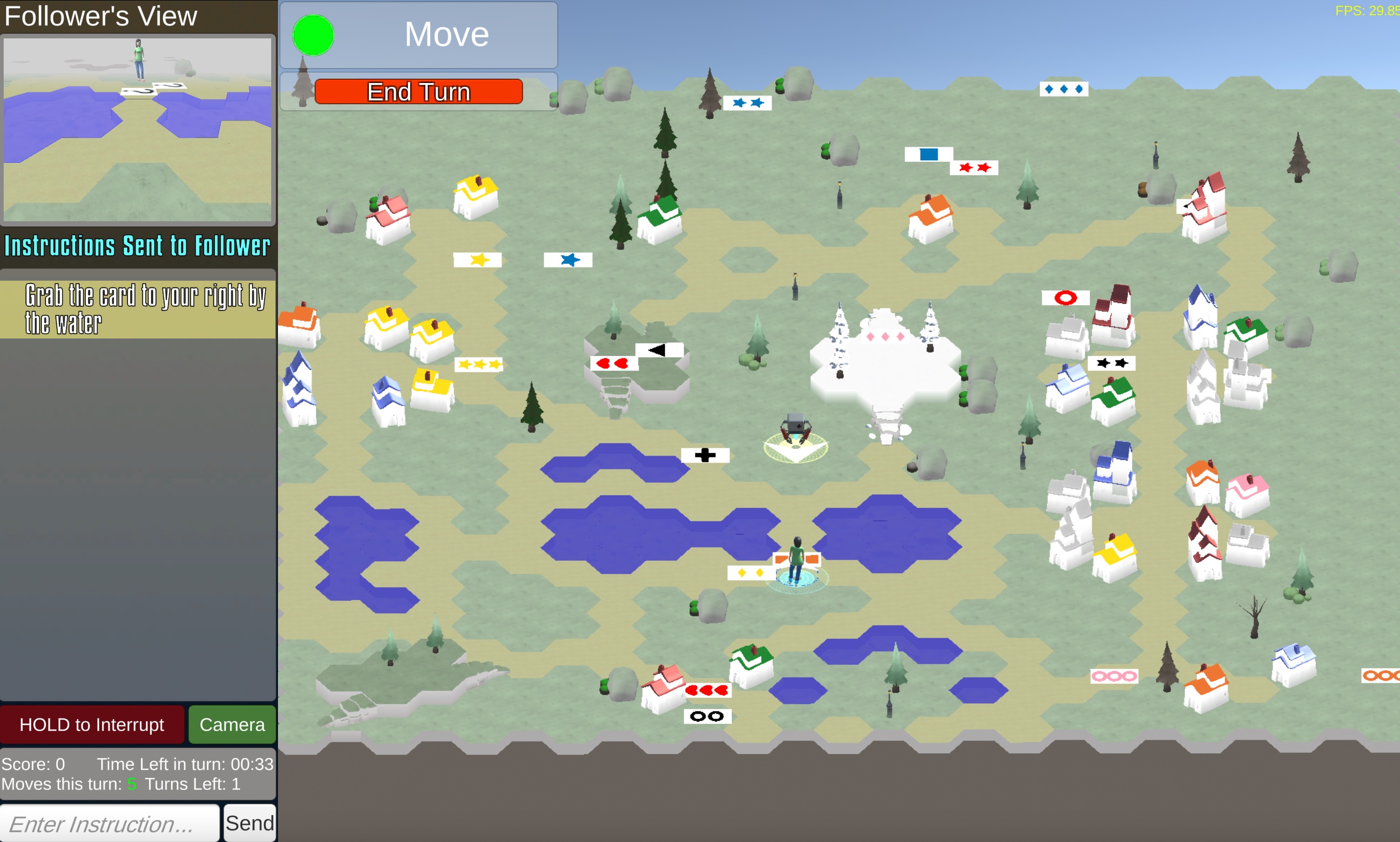}}
        \caption{An overhead view of a complete environment with the leader user interface.}\label{fig:scenario:env}        
    \end{subfigure}
    \begin{subfigure}[b]{0.2425\textwidth}
        \frame{\includegraphics[width=\linewidth,keepaspectratio,clip,trim=0 30 0 20]{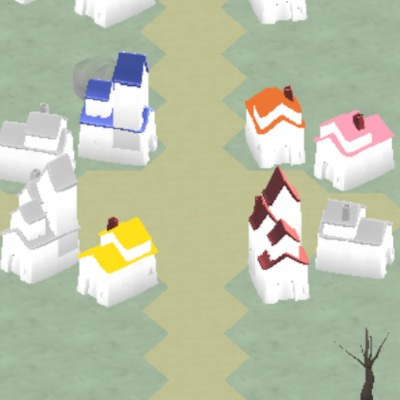}}
        \caption{A cluster of houses.}\label{fig:scenario:houses}
    \end{subfigure}
    \begin{subfigure}[b]{0.2425\textwidth}
        \frame{\includegraphics[width=\linewidth,keepaspectratio,clip,trim=0 20 0 30]{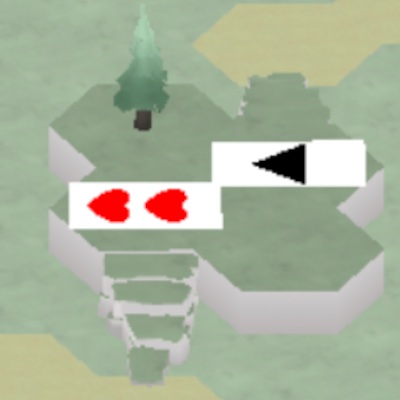}}
        \caption{A mountain with ramps.}\label{fig:scenario:moutain}
    \end{subfigure}
    \begin{subfigure}[b]{0.2425\textwidth}
        \frame{\includegraphics[width=\linewidth,keepaspectratio,clip,trim=0 20 0 30]{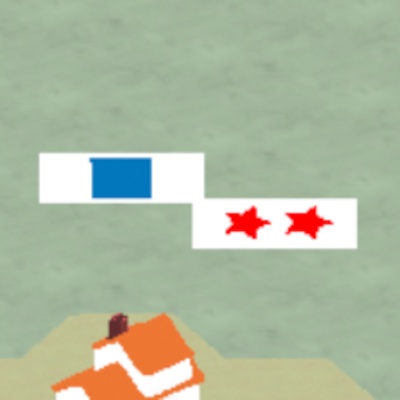}}
        \caption{Cards in the environment.}\label{fig:scenario:cards}
    \end{subfigure}
    \begin{subfigure}[b]{0.2425\textwidth}
        \frame{\includegraphics[width=\linewidth,keepaspectratio,clip,trim=0 30 0 20]{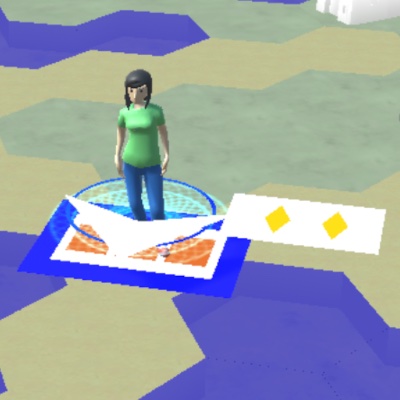}}
        \caption{The leader selecting a card.}\label{fig:scenario:cardselect}
    \end{subfigure}
    \begin{subfigure}[b]{0.245\textwidth}
        \frame{\includegraphics[width=\linewidth,keepaspectratio,clip,trim=0 15 0 4]{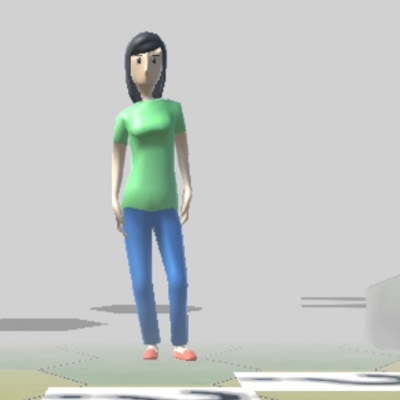}}
        \caption{The leader character.}\label{fig:scenario:leader}
    \end{subfigure}
    \begin{subfigure}[b]{0.245\textwidth}
        \frame{\includegraphics[width=\linewidth,keepaspectratio,clip,trim=0 15 0 4]{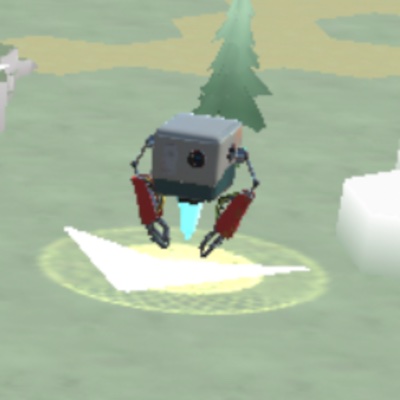}}
        \caption{The follower character.}\label{fig:scenario:follower}
    \end{subfigure}
    \begin{subfigure}[b]{0.49\textwidth}
        \frame{\includegraphics[width=\linewidth,keepaspectratio,clip,trim=0 39 0 0]{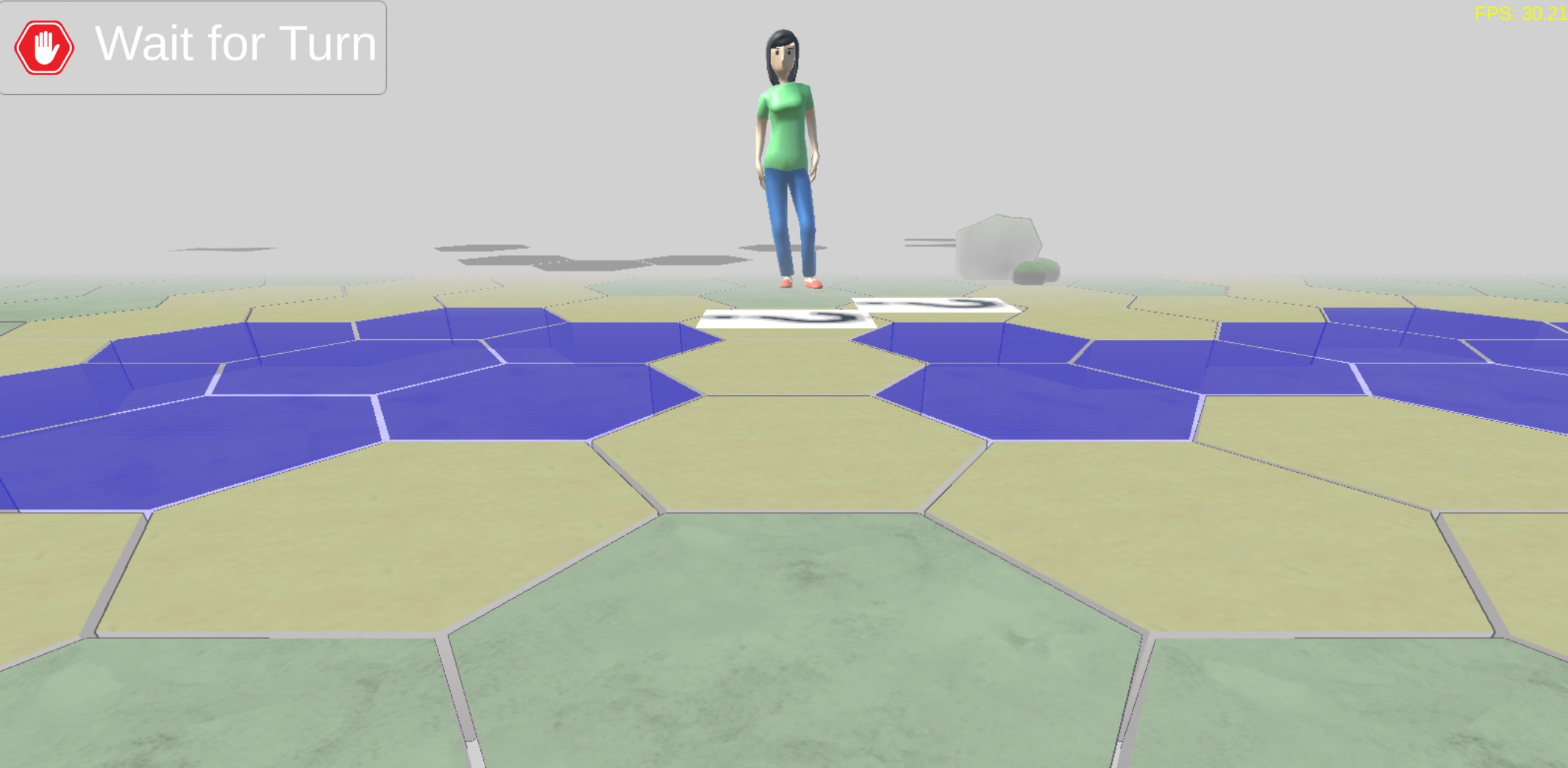}}
        \caption{The follower point of view.}\label{fig:scenario:followerpov}
    \end{subfigure}
    \begin{subfigure}[b]{\textwidth}
        \centering
        \frame{\includegraphics[width=0.49\linewidth,keepaspectratio]{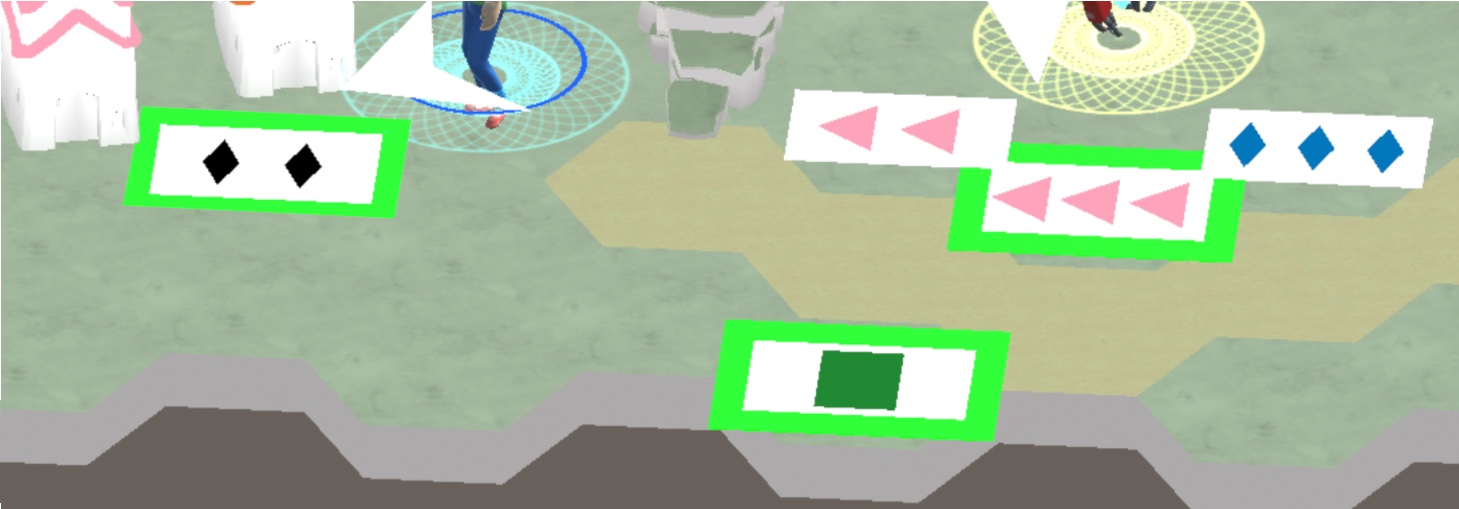}}~\frame{\includegraphics[width=0.49\linewidth,keepaspectratio]{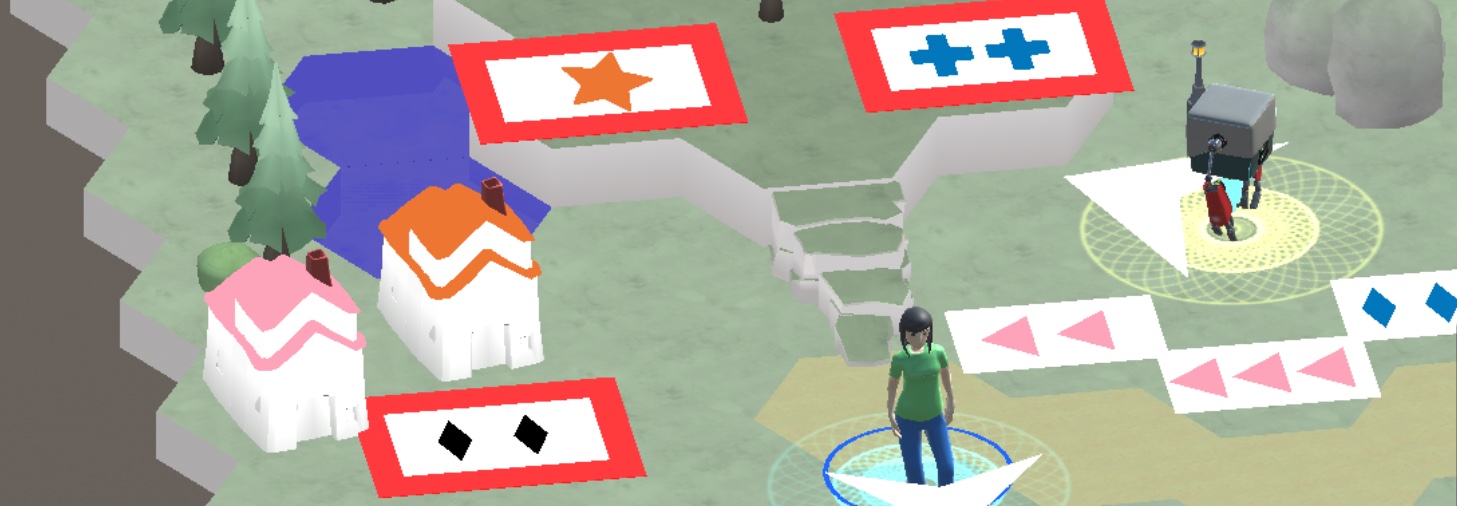}}
        \caption{Valid (left) and invalid (right) sets of selected cards.}\label{fig:scenario:cardsets}
    \end{subfigure}
    \caption{Images of the game environment and UI. All images are taken from the same environment state.}\label{fig:scenario}
    \vspace{-10pt}
\end{figure*}

\gamename largely implements the interaction scenario introduced by \cite{Suhr2019:cerealbar} in the \cerealbar environment with several modifications. The interaction takes place in a procedurally generated spatial environment and includes two agents that collaborate together to complete card collection tasks and coordinate using natural language. \autoref{fig:scenario:env} shows an instance of the environment.

The environment is a procedurally generated 3D map made of a grid of hexagons (\autoref{fig:scenario:env}). It includes lakes, mountains (\autoref{fig:scenario:moutain}), paths, open spaces, and  landmarks. 
A new environment is generated for each game. 
\gamename includes improved visuals and generation compared to \cerealbar. 
For example, \gamename map generation includes semantic biases: houses are generated to cluster together and form towns (\autoref{fig:scenario:houses}) and paths are generated to connect between meaningful areas in the map, such as towns and mountains. 
Landmark instances vary visually to elicit richer language. For example, houses are generated with different roof colors and number of floors (\autoref{fig:scenario:houses}). 
The environment also includes randomly placed cards (\autoref{fig:scenario:cards}). 
Each card shows 1--3 copies of one of a few possible shapes in one of a few possible colors. 

The interaction involves two agents, a leader (\autoref{fig:scenario:leader}) and a follower (\autoref{fig:scenario:follower}), that collaborate together to complete tasks, but differ in their observations of the environments and abilities. 
Both the leader and the follower move in the environment, by moving between neighboring hexagons or by turning in place to change orientation. 
The agents select and deselect cards by moving over them (\autoref{fig:scenario:cardselect}).

The goal of the agents is to select valid sets of cards. A valid set includes three cards, where each color, shape, and count are unique (\autoref{fig:scenario:cardsets}). 
The agents select sets together. When the currently selected cards form a valid set, they disappear, the agents together receive one point, three new randomly selected cards appear in random positions, and the agents receive additional turns. 
The number of turns added diminishes with each set completion. 
Asymmetries between the two agents make collaboration critical for success.

The leader sees a complete overhead view of the environment (\autoref{fig:scenario:env}), while the follower only sees what is ahead from a first-person view (\autoref{fig:scenario:followerpov}).  
\gamename introduces two optional observability features not present in \cerealbar. 
First, the patterns on unselected cards may be hidden from the follower, instead displaying a quesiton mark on all cards. 
Second, \gamename allows to control how far the follower sees ahead of them with a fog that is present only in the follower view. 
The observability gap means the leader is in charge of planning how the agents  operate. 
If the follower acts independently of the leader plans, the interaction will be suboptimal, because follower actions are likely to conflict with leader actions and the partial view of the environment does not allow for optimal planning of goals and movement. 

The agents move in turns, with a limited number of steps per turn. 
Each movement (forward, left, right, or backward) consumes a single step. 
Turns are time limited to keep the interaction moving and avoid long wait periods for the inactive agent. The exact time budget is customizable, but we generally provide significantly more time for the leader turns, so they can plan as needed. 
Turns alternate between the follower and leader. 
The follower has significantly more steps than the leader per turn. 
This means the follower is able to move further in each turn, and potentially accomplish much more in each turn. 
This ability gap makes it critical for the leader to collaborate with the follower, rather than ignore the follower and attempt to accomplish tasks on their own, a suboptimal strategy. 

\begin{figure*}[htbp]
    \centering
    \includesvg[width=0.98\textwidth]{diagrams/client-server-architecture.svg}
    \vspace{-10pt}
    \caption{The \gamename system architecture.}\label{fig:system}
    \vspace{-10pt}
\end{figure*}

The agents coordinate via uni-directional natural language instruction, the only form of coordination available. 
During a leader turn, in addition to moving in the environment, the leader can send text instructions to the follower. 
The follower executes the leader instructions and indicates when an instruction is complete. 
The leader can queue multiple instructions, but the follower only sees past instructions and the one they are currently executing. 
Because the follower does not see future instructions, alignment between the actions recorded and the instruction displayed is guaranteed. 
The leader can also cancel the instruction the follower is executing alongside all future instructions in the queue during the follower turn. 
This is intended to halt very bad executions, and reduce their overall cost, for example by having to correct drastic departures from the leader plan. 

Instruction writing and sending by the leader, and marking them as complete by the follower do not consume steps. 
Leaders may write as many instructions as they wish during a single turn, and followers are not taxed if the tasks are given in multiple instructions that they need to mark as complete.  
Exempting the language channel from the budget of actions per turn aims to reduce the influence of the turn systems on the language produced. 
The combination of collaboration incentives (i.e., because of the capability differences between the agents) and the exclusivity of the language channel for communication makes effective natural language instruction essential for successful interactions.\footnote{Depending on the environment configuration, it is possible for one of the agents to operate alone if the cards forming a set are really close and the other agents does not move. This can allow 1--2 set completions. A higher score without collaboration via language coordination is extremely unlikely.}

%% file: 40-impl.tex
\section{Framework Implementation}\label{sec:impl}

The \gamename framework has three main components: a Python server, a Unity client, and a Python headless client. 
The game logic is orchestrated from the server, allowing to customize the interaction without modifying Unity code. 
The Python client simplifies the interaction between learning processes and the system, for example during reinforcement learning. 
\autoref{fig:system} visualizes the architecture. 

\gamename's design emphasizes customizability, as much as possible, without modifying Unity code, a skill that is less common among researchers. 
This motivates placing the game logic on the Python server, a decision that dictates the client-server communication design. 
However, modifications that require updating the client user interface, such as adding bi-directional communication or translating the UI to other languages, do require modifying Unity coding.

\subsection{Server}\label{sec:impl:server}

The server architecture is split into modules by logical function. 
We use asynchronous coroutines to reduce latency efficiently and keep the compute needs small. 
The platform is parameterized via a configuration file that is loaded by the server. 

\paragraph{Map Generation}

 Map generation is relatively expensive compared to other processes on the server, mainly because we may use multiple search iterations for routing paths between landmarks and to prevent the leader or follower from spawning in closed-off regions. 
 We mitigate potential lag because of server load by preparing a pool of maps in advance, which we refill during idle periods. 

\paragraph{Player Lobbies}

The server supports multiple lobbies concurrently. 
Separate lobbies provide different player pairing strategies, such as for human-human and human-model games. 
Players wait in a lobby until they are paired for a game. 
Each lobby maintains multiple queues for pairing players and assigning roles according to their experience or other information. 
For example, by default, we distinguish between expert and novice players, and prioritize pairing experts as leaders with novices as followers.
Each lobby maintains active game rooms of different types,  such as for standard games, tutorials, game replays, and custom scenarios.  
Each game room contains a game state machine and websocket connections to the clients.

\paragraph{Data Storage}
Game events are recorded into an sqlite3 database, which  allows for efficient interaction with game data. Each game is represented as a linear list of events, which can be replayed to recreate game state at a particular moment in time.

\paragraph{Data Portal}

The data portal provides an interface to view game records and statistics. 
The web data browser shows game-specific recordings, including turns, instructions, and individual player actions. Each game record also includes a link to launch a game replay using the web client. 
There is also a web page with live statistics, such as the mean and median scores, and a page to download an archive of all server data. 
The data portal also provides an HTTP API for programmatic data access.

\paragraph{Map and Scenario Editor}

Maps are generated procedurally by default. 
\gamename also provides a map editor for researchers to place users in controlled scenarios. 
A real-time API allows attaching to an interaction and update the map in response to the game state, enabling dynamic interactions.

\subsection{Web Client}\label{sec:impl:webclient}

The web client is developed using the Unity game engine, and is packaged as a WebAssembly binary. The client receives game states, actions, and instructions from the server. 
We design the client to be thin, putting as much of the game logic and configuration on the server  as possible, so that changes to game logic can be made purely in Python.
We designed the gameplay user interface (UI) to be accessible and easy to learn by incorporating feedback from players. All UI elements are clustered together, and have keyboard shortcuts. \autoref{fig:scenario:env} shows the leader interface during a leader turn. 

Beyond gameplay, the web client provides a tutorial to onboard players to the game by stepping them through a game interaction accompanied by prompts and tooltips. 
The tutorial flow is specified on the server, and can be modified easily. 
For example, rephrasing the tutorial instructions or translating them to other languages is relatively simple and can be achieved by updating the specifications in the server code. 
The web client also provides game replay, which is activated by adding URL parameters when the HTML page is loaded. The parameters are added automatically to links in the web data browser (\autoref{sec:impl:server}).

\subsection{Python Client}

The programmatic Python client API supports fast lightweight interaction with the game. 
It is designed for machine learning processes that require interacting with the game environment, such as reinforcement learning~\cite{Sutton1998:rl-book-second}, and can be used to deploy agents interacting with human players or agent-agent interactions.  
Interaction through this API are similar to interactions with the Unity client, except that recording is optional to reduce overhead. 
When recorded, they can be replayed using the Unity client. 
We also provide an OpenAI Gym-style wrapper for the Python API.
\autoref{fig:game_loop} shows example code.

\begin{figure}[t]
    \centering
    \begin{minted}[
frame=lines,
framesep=2mm,
baselinestretch=1.2,
bgcolor=black,
fontsize=\scriptsize
]{python}
def PlayGameAsFollower(game):
    game_state = game.initial_state()
    # The game starts with the leader's turn. 
    # Wait for follower's turn by executing a noop.
    game_state = game.step(Action.NoopAction())
    while not game.over():
        action = get_action(game_state)
        game_state = game.step(action)
    (_, _, turn_state, _, _, _) = game_state
    print(f"Game over. Score: {turn_state.score}")
    \end{minted}
    \caption{Example code  using the Python API.}\label{fig:game_loop}
    \vspace{-10pt}
\end{figure}

%% file: 50-tasks.tex
\section{Example Task Formulations}\label{sec:tasks}

\gamename is well suited to study a variety of tasks, with emphasis on learning and evaluation in collaborative interactions with human agents, such as:

\paragraph{Instruction Following}

The task of instruction following is to map a start state observation from the follower perspective and a leader instruction to a sequence of actions. 
After each action, the agent receives a new observation. 
\citet{Suhr2019:cerealbar} studied this problem with \cerealbar by learning from recorded human-human interactions, and \citet{Suhr2022:continualfollowing} studied it within a continual learning from human feedback scenario. Both approaches were evaluated by deploying follower agents to interact with human leaders.

\paragraph{Instruction Generation}

The task of instruction generation is to generate a leader instruction for the follower to execute given an observation of the world state from the leader perspective. This requires planning the cards the two agents should select, divide the tasks, plan trajectories, and express the intended follower trajectory in a natural language instruction. 
\citet{Kojima2021:gen-learn} focused on the problem of mapping deterministically generated plans to natural language instructions, and proposed a continual learning approach for learning by observing human follower behavior.

\paragraph{Emergent Communication}

\gamename is particularly well suited to study emergent communication in multi-agent systems~\cite{Lazaridou2020:emegent-lang-survey}. 
The goal is to jointly learn separate models for the leader and follower. 
The two models generate actions to move in the world. The leader model additionally generates instructions, which the follower model is conditioned on. The learning can be driven by performance in the game. 
\gamename easily allows to integrate human agents into the learning and evaluation processes, bringing natural human language into the process. 
Alternating between interaction between agent-agent and agent-human interactions has the potential to address the language drift problem~\cite{Lee2019:countering-lang-drift}.

%% file: 60-crowdsourcing.tex
\section{Crowdsourcing Process}\label{sec:crowdsourcing}

\input{tables/game_stats.tex}

\gamename poses several relatively demanding crowdsourcing tasks. 
Human-human interactions require pairing two workers for real-time play over extended time. 
We design a process to collect \gamename interactions via crowdsourcing, either for games where both roles are controlled by human players, or where one of the sides is controlled by a learned model. 
The task-focused design of \gamename naturally allows an effective incentive structure by tying game performance with compensation.

The key to our process is gradual training of workers. 
A new worker first starts with a tutorial and a qualifier quiz that covers the relatively simple role of the follower. 
The follower role requires following the leader instructions by controlling the character in the game. 
The worker is then qualified to the follower role only, and is paired by joining a dedicated follower-only queue in the lobby. 
Focusing on the follower role only simplifies the learning curve, and much of the learning required for the leader role takes place on the job, as the worker collaborates with more experienced leaders.

Once the worker displays sufficient level of performance for several games, they are invited to qualify as a leader by taking a leader tutorial and a quiz. 
The second tutorial is both longer and more complex than the follower tutorial, and includes both planning and instruction writing. 
Once the worker completes the tutorial and passes the quiz, they are qualified to the leader role, and can then participate in tasks as both leader or follower. 

We design the lobby to pair workers based on experience. 
Because the leader role is significantly more critical to the effectiveness of the interaction and the quality of language data, we prioritize workers with better performance for it. 
We measure worker performance, keeping track of the mean game score in the most recent games. If two leader-qualified players are waiting in the lobby for matching, we will assign the leader role to the higher performing of the two. 

The pay structure includes a base pay for connecting to the game, and an increasing bonus for each point. 
Both workers, the leader and follower, get the base pay and the additional bonus per point, tightly connecting compensation to their collaborative performance.  
Because the leader role is more complex, we provide an additional relative bonus to the worker in the leader role.

%% file: tables/game_stats.tex
\begin{table*}[t]
    \centering
    \footnotesize
    \begin{tabular}{@{}lccccc@{}}
        \toprule
        \textbf{Dataset} & \textbf{\# Games} & \textbf{\# Instructions} & \textbf{Mean Score} & \textbf{Vocabulary} & \textbf{Mean Instruction Length}\\ 
        \midrule
        Training Data & 185 & 3{,}439 & 6.42 $\pm$ 4.88 & 714 & 10.95 $\pm$ 5.29  \\
        Human-Human Deployment & 187 & 3{,}404 & 6.69 $\pm$ 4.51 & 728 & 11.73 $\pm$ 6.09 \\
        Human-Model Deployment & 188 & 2{,}869 & 3.15 $\pm$ 3.29 & 542 & 9.62 $\pm$ 5.28 \\ 
        \bottomrule
    \end{tabular}
    \vspace{-5pt}
    \caption{Data and interaction statistics for the human-human training data, and the two side-by-side deployments.}\label{tab:game_stats}
    \vspace{-10pt}
\end{table*}

%% file: 70-data.tex
\section{CB2 Demonstration Deployment}\label{sec:deployment}

We demonstrate the functionality and potential of \gamename via deployment, including collecting a corpus of human-human interaction that we release, training a follower baseline model, and evaluating it in interaction with human leaders.

\paragraph{Human Games Data}
We follow the crowdsourcing process outlined in Section \ref{sec:crowdsourcing} to collect games between human leaders and followers. 
We collect 185 games containing 3{,}439 instructions.
\autoref{tab:game_stats} provides data statistics.

\paragraph{Model and Learning}
We train an instruction following model with a behavior cloning objective using the collected human-human data. 
We filter out poor games to improve training data quality, applying heuristics such as removing games where over $20\%$ of instructions are cancelled. 
Our model architecture is based on the Decision Transformer~\cite{chen2021decision, putterman2022pretraining}. 
Follower observations are embedded using \textsc{HexaConv}~\cite{hoogeboom2018hexaconv} because of the hexagonal structure of the map. 
The observations are centered on the follower's position and rotated such that the follower is always facing the same direction. 
This baseline model conditions only on the current instruction for simplicity, similar to the model in \citet{Suhr2019:cerealbar}. In contrast though, it does not assume full observability.

\paragraph{Results}
We deploy our baseline model as a system demonstration on Amazon Mechanical Turk. 
We evaluate it with 188 human-model interactions, conducted side-by-side in a randomized experiment with 187 human-human interactions. Human leaders are told that they can be matched with a human or a bot follower in the task description, but are not made aware of who they are interacting with in a specific interaction.
\autoref{tab:game_stats} shows data and interaction statistics for our training data and final deployments. 
Overall, our models enable effective human-model collaboration in \gamename, but at significantly lower performance than observed in human-human games. 
This is similar to the results of \citet{Suhr2019:cerealbar}, although the numbers are not comparable because of the different environment. 

Human leaders were able to infer relatively consistently the type of their partner in each interaction. 
This is indicated by differences in the human leader behavior when comparing human-human and human-model interactions. 
For instance, the vocabulary human leaders use in interactions with the model is smaller compared to when interacting with human followers and the instructions are shorter. 
Qualitatively, we observe that instructions in human-human interactions more often use exclamations (e.g., ``oh,'' ``shoot,'' and ``oops'') and informal speech, with abbreviations such as ``btw'' and ``lol'' or words such as ``chill'' and ``kay.'' 
We also found that human leaders in human-human games tend to praise their partners, with words such as ``awesome,'' ``wonderful,'' ``perfect'' or ``great'' appearing uniquely in instructions from human-human games. 
The difference is also seen in game statistics. 
For instance, $16.54\%$ and $12.70\%$ of the times followers and leaders selected a card in human-model games, it was to deselect an already selected card, compared to $8.68\%$ and $8.78\%$ for human-human games. Our results illustrate the challenge posed by \gamename, and the importance of the kind of deployment \gamename enables.

%% file: 80-conclusion.tex
\section{Conclusion}\label{sec:conclusion}

\gamename is a multi-agent research platform to study natural language instruction in collaborative, embodied environments. 
A core objective of \gamename is to enable scaleable studies where human agents interact with learned models, potentially over long periods of time. 
\gamename is designed to be easy to use and customize, with emphasis on accessibility for researchers with limited game development experience. 
It is designed from the ground up for machine learning, and includes a headless fast Python client API to support learning processes and to deploy learned models to interact with human users.

%% file: 90-acks.tex
\section*{Acknowledgements}

This research was supported by NSF under grant No. 1750499, ARO W911NF21-1-0106. 
We thank Alane Suhr and Noriyuki Kojima for technical discussions and utility code, and the participating MTurk workers for their work and feedback.

%% file: 95-ethics.tex
\section*{Ethical Considerations}\label{sec:ethics}

\gamename is a research environment. 
The focus on a relatively restricted 3D environment reduces the potential for ethical risks. 
Our use of \gamename has received exemption status by our institution's IRB office. We recommend that researchers using \gamename obtain IRB approval or exemption for their studies from their institution's IRB office, or an equivalent body. 
More broadly, systems that learn from interaction with users raise risks of adopting negative behavior patterns from their users. 
This is especially an issue in certain contexts, such as open ended conversational interfaces or chatbots. 
This is an important direction for future work. \gamename can be used to study such adversarial usage scenarios in a relatively safe way.

%% file: 00-cb2demo.bbl
\begin{thebibliography}{30}
\expandafter\ifx\csname natexlab\endcsname\relax\def\natexlab#1{#1}\fi

\bibitem[{Anderson et~al.(2018)Anderson, Wu, Teney, Bruce, Johnson,
  S{\"u}nderhauf, Reid, Gould, and van~den Hengel}]{Anderson:18r2r}
Peter Anderson, Qi~Wu, Damien Teney, Jake Bruce, Mark Johnson, Niko
  S{\"u}nderhauf, Ian Reid, Stephen Gould, and Anton van~den Hengel. 2018.
\newblock Vision-and-language navigation: Interpreting visually-grounded
  navigation instructions in real environments.
\newblock In \emph{The IEEE Conference on Computer Vision and Pattern
  Recognition}.

\bibitem[{Andreas et~al.(2017)Andreas, Dragan, and Klein}]{Andreas:17}
Jacob Andreas, Anca Dragan, and Dan Klein. 2017.
\newblock \href {https://doi.org/10.18653/v1/P17-1022} {Translating neuralese}.
\newblock In \emph{Proceedings of the Annual Meeting of the Association for
  Computational Linguistics}.

\bibitem[{Artzi and Zettlemoyer(2013)}]{Artzi:13}
Yoav Artzi and Luke Zettlemoyer. 2013.
\newblock \href {http://aclweb.org/anthology/Q13-1005} {Weakly supervised
  learning of semantic parsers for mapping instructions to actions}.
\newblock \emph{Transactions of the Association of Computational Linguistics},
  1.

\bibitem[{Blukis et~al.(2018)Blukis, Brukhim, Bennett, Knepper, and
  Artzi}]{Blukis:18drone}
Valts Blukis, Nataly Brukhim, Andrew Bennett, Ross~A. Knepper, and Yoav Artzi.
  2018.
\newblock Following high-level navigation instructions on a simulated
  quadcopter with imitation learning.
\newblock In \emph{Proceedings of the Robotics: Science and Systems
  Conference}.

\bibitem[{Chen and Mooney(2011)}]{Chen:11}
David~L. Chen and Raymond~J. Mooney. 2011.
\newblock Learning to interpret natural language navigation instructions from
  observations.
\newblock In \emph{Proceedings of the National Conference on Artificial
  Intelligence}.

\bibitem[{Chen et~al.(2021)Chen, Lu, Rajeswaran, Lee, Grover, Laskin, Abbeel,
  Srinivas, and Mordatch}]{chen2021decision}
Lili Chen, Kevin Lu, Aravind Rajeswaran, Kimin Lee, Aditya Grover, Misha
  Laskin, Pieter Abbeel, Aravind Srinivas, and Igor Mordatch. 2021.
\newblock Decision transformer: Reinforcement learning via sequence modeling.
\newblock \emph{Advances in Neural Information Processing Systems}.

\bibitem[{Daniele et~al.(2016)Daniele, Bansal, and Walter}]{daniele2016natural}
Andrea~F Daniele, Mohit Bansal, and Matthew~R Walter. 2016.
\newblock Natural language generation in the context of providing indoor route
  instructions.
\newblock In \emph{Proceedings Robotics: Science and Systems Workshop on Model
  Learning for Human-Robot Communication}.

\bibitem[{Djalali et~al.(2012)Djalali, Lauer, and
  Potts}]{Djalali12:cards-preference}
Alex Djalali, Sven Lauer, and Christopher Potts. 2012.
\newblock Corpus evidence for preference-driven interpretation.
\newblock In \emph{Logic, Language and Meaning}.

\bibitem[{Effenberger et~al.(2021)Effenberger, Singh, Yan, Suhr, and
  Artzi}]{Effenberger2021:cerealbar-analysis}
Anna Effenberger, Rhia Singh, Eva Yan, Alane Suhr, and Yoav Artzi. 2021.
\newblock Analysis of language change in collaborative instruction following.
\newblock In \emph{Findings of the Association for Computational Linguistics:
  EMNLP}.

\bibitem[{Fried et~al.(2018)Fried, Andreas, and
  Klein}]{Fried:17pragmatic-models}
Daniel Fried, Jacob Andreas, and Dan Klein. 2018.
\newblock \href {https://doi.org/10.18653/v1/N18-1177} {Unified pragmatic
  models for generating and following instructions}.
\newblock In \emph{Proceedings of the Conference of the North American Chapter
  of the Association for Computational Linguistics: Human Language
  Technologies}.

\bibitem[{Harnad(1990)}]{Harnad1990:symbol-grounding-problem}
Stevan Harnad. 1990.
\newblock The symbol grounding problem.
\newblock \emph{Physica D: Nonlinear Phenomena}, 42.

\bibitem[{Hoogeboom et~al.(2018)Hoogeboom, Peters, Cohen, and
  Welling}]{hoogeboom2018hexaconv}
Emiel Hoogeboom, Jorn~W.T. Peters, Taco~S. Cohen, and Max Welling. 2018.
\newblock \href {https://openreview.net/forum?id=r1vuQG-CW} {Hexaconv}.
\newblock In \emph{International Conference on Learning Representations}.

\bibitem[{Jayannavar et~al.(2020)Jayannavar, Narayan-Chen, and
  Hockenmaier}]{Jayannavar2020:instructions-minecraft-dialogue}
Prashant Jayannavar, Anjali Narayan-Chen, and Julia Hockenmaier. 2020.
\newblock \href {https://doi.org/10.18653/v1/2020.acl-main.232} {Learning to
  execute instructions in a {M}inecraft dialogue}.
\newblock In \emph{Proceedings of the Annual Meeting of the Association for
  Computational Linguistics}.

\bibitem[{Kiseleva et~al.(2022)Kiseleva, Skrynnik, Zholus, Mohanty, Arabzadeh,
  C{\^o}t{\'e}, Aliannejadi, Teruel, Li, Burtsev et~al.}]{kiseleva2022iglu}
Julia Kiseleva, Alexey Skrynnik, Artem Zholus, Shrestha Mohanty, Negar
  Arabzadeh, Marc-Alexandre C{\^o}t{\'e}, Mohammad Aliannejadi, Milagro Teruel,
  Ziming Li, Mikhail Burtsev, et~al. 2022.
\newblock Iglu 2022: Interactive grounded language understanding in a
  collaborative environment at neurips 2022.
\newblock \emph{arXiv preprint arXiv:2205.13771}.

\bibitem[{Kojima et~al.(2021)Kojima, Suhr, and Artzi}]{Kojima2021:gen-learn}
Noriyuki Kojima, Alane Suhr, and Yoav Artzi. 2021.
\newblock \href {https://doi.org/10.1162/tacl_a_00428} {Continual learning for
  grounded instruction generation by observing human following behavior}.
\newblock \emph{Transactions of the Association for Computational Linguistics},
  9.

\bibitem[{Ku et~al.(2020)Ku, Anderson, Patel, Ie, and
  Baldridge}]{Ku2020:room-across-room}
Alexander Ku, Peter Anderson, Roma Patel, Eugene Ie, and Jason Baldridge. 2020.
\newblock \href {https://doi.org/10.18653/v1/2020.emnlp-main.356}
  {Room-across-room: Multilingual vision-and-language navigation with dense
  spatiotemporal grounding}.
\newblock In \emph{Proceedings of the Conference on Empirical Methods in
  Natural Language Processing}.

\bibitem[{Lazaridou and Baroni(2020)}]{Lazaridou2020:emegent-lang-survey}
Angeliki Lazaridou and Marco Baroni. 2020.
\newblock Emergent multi-agent communication in the deep learning era.
\newblock \emph{ArXiv}, abs/2006.02419.

\bibitem[{Lazaridou et~al.(2017)Lazaridou, Peysakhovich, and
  Baroni}]{Lazaridou:17}
Angeliki Lazaridou, Alexander Peysakhovich, and Marco Baroni. 2017.
\newblock Multi-agent cooperation and the emergence of (natural) language.
\newblock In \emph{International Conference on Learning Representations}.

\bibitem[{Lee et~al.(2019)Lee, Cho, and Kiela}]{Lee2019:countering-lang-drift}
Jason Lee, Kyunghyun Cho, and Douwe Kiela. 2019.
\newblock Countering language drift via visual grounding.
\newblock In \emph{Proceedings of the Conference on Empirical Methods in
  Natural Language Processing}.

\bibitem[{MacMahon et~al.(2006)MacMahon, Stankiewics, and
  Kuipers}]{MacMahon:06}
Matthew MacMahon, Brian Stankiewics, and Benjamin Kuipers. 2006.
\newblock Walk the talk: Connecting language, knowledge, action in route
  instructions.
\newblock In \emph{Proceedings of the National Conference on Artificial
  Intelligence}.

\bibitem[{Mei et~al.(2016)Mei, Bansal, and Walter}]{Mei:16generation}
Hongyuan Mei, Mohit Bansal, and R.~Matthew Walter. 2016.
\newblock \href {https://doi.org/10.18653/v1/N16-1086} {What to talk about and
  how? {S}elective generation using lstms with coarse-to-fine alignment}.
\newblock In \emph{Proceedings of the Conference of the North American Chapter
  of the Association for Computational Linguistics: Human Language
  Technologies}.

\bibitem[{Misra et~al.(2017)Misra, Langford, and Artzi}]{Misra:17instructions}
Dipendra Misra, John Langford, and Yoav Artzi. 2017.
\newblock Mapping instructions and visual observations to actions with
  reinforcement learning.
\newblock In \emph{Proceedings of the Conference on Empirical Methods in
  Natural Language Processing}.

\bibitem[{Narayan-Chen et~al.(2019)Narayan-Chen, Jayannavar, and
  Hockenmaier}]{Narayan2019:collaborative-minecraft-diaglogue}
Anjali Narayan-Chen, Prashant Jayannavar, and Julia Hockenmaier. 2019.
\newblock \href {https://doi.org/10.18653/v1/P19-1537} {Collaborative dialogue
  in {M}inecraft}.
\newblock In \emph{Proceedings of the Annual Meeting of the Association for
  Computational Linguistics}.

\bibitem[{Potts(2012)}]{Potts:12}
Christopher Potts. 2012.
\newblock Goal-driven answers in the {C}ards dialogue corpus.
\newblock In \emph{Proceedings of the West Coast Conference on Formal
  Linguistics}.

\bibitem[{Putterman et~al.(2022)Putterman, Lu, Mordatch, and
  Abbeel}]{putterman2022pretraining}
Aaron~L Putterman, Kevin Lu, Igor Mordatch, and Pieter Abbeel. 2022.
\newblock \href {https://openreview.net/forum?id=eCPCn25gat} {Pretraining for
  language conditioned imitation with transformers}.
\newblock \emph{Offline Reinforcement Learning Workshop at Neural Information
  Processing Systems}.

\bibitem[{Shridhar et~al.(2020)Shridhar, Thomason, Gordon, Bisk, Han, Mottaghi,
  Zettlemoyer, and Fox}]{Shridhar2020:alfred}
Mohit Shridhar, Jesse Thomason, Daniel Gordon, Yonatan Bisk, Winson Han,
  Roozbeh Mottaghi, Luke Zettlemoyer, and Dieter Fox. 2020.
\newblock {ALFRED}: A benchmark for interpreting grounded instructions for
  everyday tasks.
\newblock In \emph{IEEE Conference on Computer Vision and Pattern Recognition}.

\bibitem[{Suhr and Artzi(2022)}]{Suhr2022:continualfollowing}
Alane Suhr and Yoav Artzi. 2022.
\newblock Continual learning for instruction following from realtime feedback.
\newblock \emph{arXiv preprint arXiv:2212.09710}.

\bibitem[{Suhr et~al.(2019)Suhr, Yan, Schluger, Yu, Khader, Mouallem, Zhang,
  and Artzi}]{Suhr2019:cerealbar}
Alane Suhr, Claudia Yan, Jack Schluger, Stanley Yu, Hadi Khader, Marwa
  Mouallem, Iris Zhang, and Yoav Artzi. 2019.
\newblock \href {https://doi.org/10.18653/v1/D19-1218} {Executing instructions
  in situated collaborative interactions}.
\newblock In \emph{Proceedings of the Conference on Empirical Methods in
  Natural Language Processing}.

\bibitem[{Sutton and Barto(1998)}]{Sutton1998:rl-book-second}
Richard~S. Sutton and Andrew~G. Barto. 1998.
\newblock \emph{Reinforcement learning: An introduction}.
\newblock MIT press.

\bibitem[{Wang et~al.(2021)Wang, Montgomery, Orbay, Birodkar, Faust, Gur,
  Jaques, Waters, Baldridge, and Anderson}]{Wang2021:generatingInstructions}
Su~Wang, Ceslee Montgomery, Jordi Orbay, Vighnesh Birodkar, Aleksandra Faust,
  Izzeddin Gur, Natasha Jaques, Austin Waters, Jason Baldridge, and Peter
  Anderson. 2021.
\newblock Less is more: Generating grounded navigation instructions from
  landmarks.
\newblock In \emph{Proceedings of the Conference on Computer Vision and Pattern
  Recognition}.

\end{thebibliography}
